# Predicting human motion intention for pHRI assistive control

Paolo Franceschi[1,2], Fabio Bertini[3], Francesco Braghin[3], Loris Roveda[4], Nicola Pedrocchi[1], and Manuel Beschi[2]

*Abstract*— This work addresses human intention identification during physical Human-Robot Interaction (pHRI) tasks to include this information in an assistive controller. To this purpose, human intention is defined as the desired trajectory that the human wants to follow over a finite rolling prediction horizon so that the robot can assist in pursuing it. This work investigates a Recurrent Neural Network (RNN), specifically, Long-Short Term Memory (LSTM) cascaded with a Fully Connected layer. In particular, we propose an iterative training procedure to adapt the model. Such an iterative procedure is powerful in reducing the prediction error. Still, it has the drawback that it is time-consuming and does not generalize to different users or different co-manipulated objects. To overcome this issue, Transfer Learning (TL) adapts the pre-trained model to new trajectories, users, and co-manipulated objects by freezing the LSTM layer and fine-tuning the last FC layer, which makes the procedure faster. Experiments show that the iterative procedure adapts the model and reduces prediction error. Experiments also show that TL adapts to different users and to the co-manipulation of a large object. Finally, to check the utility of adopting the proposed method, we compare the proposed controller enhanced by the intention prediction with the other two standard controllers of pHRI.

## I. INTRODUCTION

Human-Robot Interaction (HRI) defines the study of safe and natural interaction between humans and robots [1]. Particularly, in modern factories, the need for collaborative manufacturing tasks shared between humans and robots has been pushing the research in this field [2]. Among the various levels of HRI, we are dealing with physical Human-Robot Interaction (pHRI) [3]. Various pHRI controllers exist to assist human operators in performing shared tasks, mainly based on the Impedance/Admittance Control and its adaptive version based on different techniques, as in [4], [5], [6]. Among the various control schema, Game-Theory (GT) based controllers are nowadays under investigation thanks to their capability to model the interaction between multiple players and provide them with "optimal" control actions. Examples of GT-based pHRI controllers are [7], [8], [9]. GT controllers consist of each player knowing the opponents' objective, which can be formulated as minimizing a cost function. Such a cost function also includes a term related to the player's intention, which in pHRI can be seen as the desire to follow a precise trajectory. Within this framework, intent detection can be defined as the need for the robot to know at least partially the human's planned action to assist toward achieving it [10].

A standard approach is to model human interaction with robots using the impedance model [11], [12]. These approaches must estimate the time-varying impedance parameters, and the model's output depends on these parameters and the specific task considered. Hence the utility of having a model comes at the cost of low flexibility. A different approach is to model the human as a feedback controller [13], [14], still requiring a control model and parameter estimation.

To overcome the modeling and parameter estimations, Machine Learning is gaining popularity in addressing such problems [15], [16]. Neural networks achieve excellent results in approximating complex non-linear systems with high uncertainties. Among the various models, Recurrent Neural Networks (RNN) are very effective when dealing with sequences. An advantage of RNNs is that they consider previous events allowing information to persist. Among the various RNN types, Long-Short Term Memory (LSTM) [17] architecture outperforms the classical RNN and is now widely adopted to solve multiple problems where the sequential pattern of the data may store important information, such as speech recognition. The same reasoning applies to human intent estimation since the previous motion state may be a manifestation of the intent.

RNN architectures have demonstrated effectiveness in the prediction of human motion. Typical approaches exploit vision-based data, as in [18],[19], and [20]. They use different RNN models to forecast the human operator's motion trajectory. The architectures used are LSTM, and Gated Recurrent Unitis (GRU), an architecture similar to LSTM. A different model is presented in [21], where the autoregressive integrated moving average (ARIMA) model uses the visual identification of the elbow motion of a human. All these works use information from the camera to monitor the human positions and predict the next steps. The main drawback is that they require specific hardware, and image data increase the duration of the training phase due to the high complexity.

In [22], FMG sensor provides the LSTM with input data.

[1]Paolo Franceschi and Nicola Pedrocchi are with the Institute of Intelligent Industrial Technologies and Systems for Advanced Manufacturing of the National Research Council of Italy (CNR-STIIMA), via Alfonso Corti 12, 20133, Milano, Italy. {*name.surname*}@stiima.cnr.it

[2]Paolo Franceschi and Manuel Beschi are with the Università di Brescia, Dipartimento di Ingegneria Meccanica ed Industriale, via Branze 38, 25123, Brescia, Italy manuel.beschi@unibs.it

[3]Fabio Bertini and Francesco Braghin are with the Department of Mechanical Engineering, Politecnico di Milano, Milano 20133, Italy francesco.braghin@polimi.it

[4]Loris Roveda is with Istituto Dalle Molle di Studi sull'Intelligenza Artificiale (IDSIA), Scuola Universitaria Professionale della Svizzera Italiana (SUPSI), Universita della Svizzera Italiana (USI), Lugano 6962, Switzerland loris.roveda@idsia.ch

This works uses multiple subjects for training but does not show any generalization of the approach. A different approach exploits Gaussian processes [23], [24]. The first proposes identifying human motion intention interacting with an exoskeleton via a sparse Gaussian process. The second models the human arm as 7 degrees of freedom (DoFs) impedance and estimates the intention as the human force by Gaussian Process. In [25], authors propose an NN-classifier that predicts the intended direction of human movement by utilizing electromyography (EMG) signals acquired from human arm muscles. This work proposes motion classification to define direction, and the robot assists but does not predict future motion intentions. In [26], an Adaptive Neural Network estimates the joint coordinates of the human lower limb interacting with an exoskeleton for rehabilitation. In contrast, [27] predict human motion intention by online learning without training. In [28], the human intent is a reference position obtained by double integration of the estimated acceleration. In [29], the model is based on Radial Basis Function Neural Networks (RBFNN). In this work, an updating law adjusts the NN weights online to guarantee estimation accuracy even when human motion intention changes. Still, the prediction horizon is one step. Interestingly, [30] and [31], propose an LSTM to predict the reference set-point at the next step but do not address any adaptation to new users or objects.

The reviewed works present relevant results but miss fundamental aspects of human intention modeling. First, using a model trained on data collected without the robot's assistance and then using it to compute robot assistance changes the robot's behavior. Therefore, the model is no more completely reliable. Moreover, predicting only one step might be helpful for assistance. Still, longer-term predictions allow Model Predictive Control (MPC) implementation and the detection of possible dangerous situations in advance (joint limits, obstacle collisions, etc.). Finally, some of the presented works rely on a model trained on different people with many data. Still, they do not address the model's adaptability to new users or situations.

This work proposes a procedure for training a model capable of estimating the desired human trajectory over a finite prediction horizon. The second objective is to keep the training time and computational resources low. In particular, this work aims at reducing the time when model adaptation is required to address new users or new situations, such as different trajectories or co-manipulated objects.

We address these two points by proposing an LSTM cascaded with Fully Connected (FC) layers (LSTM+FC). First, an iterative training procedure allows adapting the model to improve the prediction error and provide proper assistance. After that, transfer learning is proposed to address the time issue related to handling new situations.

## II. METHOD

This section briefly introduces the cooperative controller to show how the proposed RNN+FC model is integrated. After that, it explains how the model is built and trained. Finally, Transfer learning of the proposed model to other users/objects is presented.

### A. Cooperative controller

This subsection briefly recalls the GT-based control modeling proposed in [9]. The motion of the robot tip is described as a Cartesian Impedance control as follows:

$$M_i \left( \ddot{x} - \ddot{x}_0 \right) + C_i \left( \dot{x} - \dot{x}_0 \right) + K_i \left( x - x_0 \right) = u_h(t) + u_r(t), \quad (1)$$

where $M_i$, $C_i$ and $K_i \in \mathbb{R}^{6\times 6}$ are the desired inertia, damping, and stiffness matrices, respectively, $\ddot{x}(t)$, $\dot{x}(t)$ and $x(t) \in \mathbb{R}^6$ are the Cartesian accelerations, velocities and positions at the end-effector, $x_0$ is the equilibrium position of the virtual spring, $u_h(t) \in \mathbb{R}^6$ is the human measured force, and $u_r(t) \in \mathbb{R}^6$ represents a virtual robot effort applied to the system. The Cartesian coordinates in $x$ are defined according to [32], with the vector $x = [p^T \ \theta^T]^T$ where $p^T$ are the position coordinates and $\theta^T$ the set of Euler angles[1].

The equation (1) can be rewritten in a linearized state-space formulation around the working point as

$$\dot{z} = Az + B_h u_h + B_r u_r, \quad (2)$$

where $z = [x \ \dot{x}]^T \in \mathbb{R}^{12}$ is the state space vector, $A = \begin{bmatrix} 0^{6\times 6} & I^{6\times 6} \\ -M_i^{-1}K_i & -M_i^{-1}C_i \end{bmatrix}$, $B_h^{12\times 6} = B_r^{12\times 6} = \begin{bmatrix} 0^{6\times 6} \\ M_i^{-1} \end{bmatrix}$, with $0^{6\times 6}$ denoting a $6 \times 6$ zero matrix and $I^{6\times 6}$ the Identity matrix. With kinematic inversion at the velocity level, it is possible to command the robot in the joint space, as

$$\dot{q}_{ref}(t) = J_a(q)^+ \dot{x}(t), \quad (3)$$

where $\dot{q}_{ref}(t) \in \mathbb{R}^n$ are the reference velocities in the joint space, $n$ represents the number of joints, and $J_a(q)^+$ is the pseudoinverse of the analytical Jacobian matrix. Joint positions are then computed via a simple integration. Assume $\dot{q} \simeq \dot{q}_{ref}$, considering that today's robots have great tracking performance in the frequency range excitable by the operator.

Given the system in (2), the objective is to define the assistive robot contribution $u_r$. We propose the use of Cooperative Game Theory to define the proper assistance. In the following, we present the CGT formulation of the two-players game, based on [33], with an extension to the agreement of a shared reference.

In a cooperative framework, the human and the robot are modeled as two agents, each one aiming at minimizing a quadratic cost function, defined as

$$J_i = \int_0^\infty \left[ (z - z_{ref,i})^T Q_{i,i} (z - z_{ref,i}) \right. \\ \left. + (z - z_{ref,j})^T Q_{i,j} (z - z_{ref,j}) + u_i^T R_i u_i \right] dt, \quad (4)$$

with $i,j = \{h,r\}$ indicating the human and the robot. Were $J_i$ are the costs that the human and the robot incur, $Q_{i,i}, Q_{i,j} \in \mathbb{R}^{12\times 12}$ matrices that weights the state and references and $R_i \in \mathbb{R}^{6\times 6}$ weights on the control input. By cooperating, a shared objective is defined as

$$J_c = \alpha J_h + (1-\alpha) J_r = \int_0^\infty \left(\tilde{z}^T Q_c \tilde{z} + u^T R_c u\right) dt, \quad (5)$$

with $\tilde{z} = z - z_{ref}$, where $z_{ref}$, $Q_c$ and $R_c$ must be defined, and $\alpha \in (0,1)$ represents the weight each player's cost has in the overall cost. Combining the two costs (4) into (5), after some calculations, can be obtained

$$Q_c = \alpha \left(Q_{h,h} + Q_{h,r}\right) + (1-\alpha)\left(Q_{r,h} + Q_{r,r}\right), \quad (6)$$

and

$$R_c = \begin{bmatrix} \alpha \hat{R}_h & 0^{6\times 6} \\ 0^{6\times 6} & (1-\alpha) R_r. \end{bmatrix} \quad (7)$$

---

[1]This choice assumes that the angular rotation maintains limited values in the target applications, mainly along one rotation axis.

Finally, defining $Q_h = \alpha Q_{h,h} + (1-\alpha)Q_{h,r}$ and $Q_r = \alpha Q_{r,h} + (1-\alpha)Q_{r,r}$, the reference $z_{ref}$ is a weighted composition of the human and robot references, that can be expressed as

$$z_{ref} = Q_c^{-1}(z_{ref,h} Q_h + z_{ref,r} Q_r). \quad (8)$$

With a further step, the system in (2) becomes

$$\dot{z} = Az + Bu, \quad (9)$$

with $A \in \mathbb{R}^{12\times 12}$ defined as before, $B \in \mathbb{R}^{12\times 12} = [B_h \; B_r]$ and $u = [u_h \; u_r]^T \in \mathbb{R}^{12\times 1}$.

The Linear Quadratic Differential Game problem can be finally formulated as a classical LQR problem:

$$\min_u J_c = \int_0^\infty \left(z^T Q_c z + u^T R_c u\right) dt \quad (10)$$
$$\text{s.t. } \dot{z} = Az + Bu$$
$$z(t_0) = z_0.$$

The problem in (10) has infinite solutions on the Pareto frontier, depending on the parameter $\alpha$. The choice of the parameter $\alpha$ is the solution to the so-called Bargaining problem, and different methods exist to define a value [34], [35]. In the proposed controller, the value of $\alpha$ assumes the role of assistance level. High values of $\alpha$ correspond to a robot more willing to help the human track the desired human trajectory. Therefore, we do not directly address the Bargaining problem but select a value of $\alpha = 0.8$, which experimentally has proven sufficiently high to assist. With this value, assuming $z_{ref} \simeq z_{ref,h}$ is reasonable.

In an LQ-CGT framework, the control action $u$ is defined as full-state feedback as

$$u = -K_{gt}\tilde{z} = -K_{gt}z + K_{gt}z_{ref}, \quad (11)$$

with $K_{gt} = R_c^{-1}B^T P$ and matrix $P$ solution of the Algebraic Riccati Equation (ARE)

$$0 = A^T P + PA^T - PBR_c^{-1}B^T P + Q_c.$$

Note that $u = [u_h, u_r]^T$. Therefore, the robot assistive contribution can be extracted by slicing the vector of control inputs. A schema of the proposed controller is visible in Figure 1.

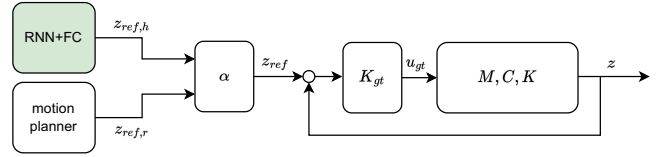

Fig. 1: The control schema of the proposed cooperative controller. The RNN+FC model proposed in this work is visible in green to identify human reference trajectory.

### B. RNN+FC

Based on (8), the robot must know the human desired reference to give proper assistance. As also discussed in [29], the following can express the human limb dynamics in interaction with a robot at the robot's tip

$$-C_h \dot{x} - K_h(x_{ref,h} - x) = u_h, \quad (12)$$

with $C_h$ and $K_h$ damping and stiffness matrices. Assuming that $C_h = C_h(x, \dot{x})$ and $K_h = K_h(x)$, the desired human motion can be defined as

$$x_{ref,h} = F(x, \dot{x}, u_h). \quad (13)$$

The function $F$ is nonlinear and time-variant. The problem becomes even more complicated if the human and the robot interact while transporting a large object, with additional inertia and different contact points. Therefore, we propose using a Recurrent Neural Network (RNN), cascaded with a Fully Connected (FC) layer (RNN+FC), to learn the complex human dynamics and provide the robot with the desired trajectory over the next horizon. In particular, among the various types, this work proposes using Long-Short Term Memory (LSTM), which has proven to have better performances for long-time series than the basic RNN.

The proposed method aims to identify and predict, over a finite rolling horizon, the desired human trajectory, given the history over a finite horizon. The RNN+FC model takes the last $k$ time instant as input and predicts the human desired trajectory over the next $N$ steps. The model accepts as inputs the actual robot positions and velocities $\mathbf{x} = [x, y, z, R, P, Y]^T$ and $\mathbf{v} = [\dot{x}, \dot{y}, \dot{z}, \omega_x, \omega_y, \omega_z]^T$, respectively, the force the human is exerting $\mathbf{u_h} = [f_x, f_y, f_z, \tau_x, \tau_y, \tau_z]^T$, and the nominal robot trajectory $\mathbf{x_{ref,r}} = [x_{ref,r}, y_{ref,r}, z_{ref,r}, R_{ref,r}, P_{ref,r}, Y_{ref,r}]^T$. Defining with $T$ the current time, the input data are defined as $\underline{\mathbf{x}} = \{\mathbf{x}_{T-k}, ..., \mathbf{x}_T\}$ and $\underline{\mathbf{v}} = \{\mathbf{v}_{T-k}, ..., \mathbf{v}_T\}$ the vector containing the positions and velocities of the past $k$ time instant, respectively, $\underline{\mathbf{u_h}} = \{\mathbf{u_h}_{T-k}, ..., \mathbf{u_h}_T\}$ the vector containing the human interaction wrench over the past $k$ time instant and $\underline{\mathbf{x_{ref,r}}} = \{\mathbf{x_{ref,r}}_{T-k}, ..., \mathbf{x_{ref,r}}_T\}$ the vector containing the reference trajectory of the robot of the past $k$ time instant. The output of the model is a finite sequence of reference positions in the horizon that goes from time $T+1$ to $T+N$, defined as $\widehat{\mathbf{x}}_{\mathbf{ref,h}} = \{\mathbf{x_{ref,h}}_{T+1}, ..., \mathbf{x_{ref,h}}_{T+N}\}$, where $\widehat{(\cdot)}$ denotes an estimate. A schema of the proposed model with inputs and output is visible in Figure 2.

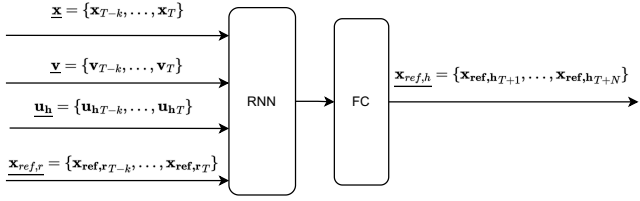

Fig. 2: Human trajectory identification using the RNN+FC.

**Algorithm 1:** The iterative training procedure
**Data:** Sample records
**Result:** Trained model $\mathcal{M}$
Record dataset $\mathcal{D}_0$ without any model
train model $\mathcal{M}_{\prime}$ with dataset $\mathcal{D}_0$
**while** $\|e_{RMS}^{k+1} - e_{RMS}^{k}\| < tol$ **do**
  Record dataset $\mathcal{D}_k$ with model $\mathcal{M}_k$;
  train model $\mathcal{M}_{k+1}$ with dataset $\mathcal{D}_k$;
**if** *new user or object* **then**
  Record dataset $\mathcal{D}_{TL}$ with model $\mathcal{M}_{k+1}$;
  train model $\mathcal{M}_{TL}$ with dataset $\mathcal{D}_{TL}$;

### C. Iterative training

The RNN+FC model prediction output is plugged into (8), allowing computing the efforts in (11). Before using the model, data have to be acquired and processed to train the input-output relationship, *i.e.* the function F in (13) in our case. Collecting data without using the model (assuming, for example, $\widehat{\mathbf{x}}_{\mathbf{ref,h}} = \underline{\mathbf{x}_{\mathbf{ref,r}}}$) allows the model training. However, once the model is being used, the robot behavior is no more the same as during the data collection phase because the assumption $\widehat{\mathbf{x}}_{\mathbf{ref,h}} = \underline{\mathbf{x}_{\mathbf{ref,r}}}$ does not hold anymore. The robot assists using that estimate. The model can predict $\mathbf{x}_{\mathbf{ref,h}}$ only with the original robot behavior based on such an assumption.

We propose an iterative training procedure to face this problem, we first collect an initial dataset $\mathcal{D}_0$ without using any model, letting $\widehat{\mathbf{x}}_{\mathbf{ref,h}} = \underline{\mathbf{x}_{\mathbf{ref,r}}}$. We train a model, call it $\mathcal{M}_0$ with the subscript $_0$ indicating it is the model trained with no model, which depends on the first dataset only $\mathcal{M}_0 = \mathcal{M}_0(\mathcal{D}_0)$. Once we have a model, a new data collection phase begins. After the second dataset $\mathcal{D}_1$ is collected, it is possible to train a second model $\mathcal{M}_1 = \mathcal{M}_1(\mathcal{M}_0, \mathcal{D}_0)$. This procedure is iterated for $K$ times until a stop criterion is reached, and the model $\mathcal{M}_K = \mathcal{M}_K(\mathcal{M}_{k-1}, \mathcal{D}_{k-1})$ is finally ready for usage. A stopping criterion can be the prediction error, defined as the average of the Root Mean Square Error (RMS),

$$e_{RMS} = \frac{1}{L}\sum_{T=1}^{L}\sqrt{\frac{1}{N}\sum_{k=T}^{T+N}(\|\widehat{\mathbf{x}}_{\mathbf{ref,h_k}} - \mathbf{x}_k\|)}^2, \quad (14)$$

where $\widehat{\mathbf{x}}_{\mathbf{ref,h_k}}$ is the predicted human intention, $\mathbf{x}_k$ the measured poses, $L$ is the length of the trajectory, and $N$ is the prediction horizon. The iterative procedure stops when $\|e_{RMS}^{k+1} - e_{RMS}^{k}\| < tol$.

### D. Transfer learning

Equation (12) differs from human to human. It describes human dynamics when the human is grasping the robot's tip but does not consider possible co-carried large objects that can be grasped from one side by the robot and from the other by the human, making the assumption invalid. The model $\mathcal{M}_K$ obtained after the iterative procedure might require additional training to adapt to new users and objects.

The main drawback of iterative procedures is they are time-consuming, requiring time for data collecting and model training. Transfer learning is a widely adopted method to speed up training starting from a pre-trained model.

Among the different TL techniques, we propose adopting a widely used strategy in computer vision or NLP domains, which consists of "freezing" some model layers and re-train on new data only a few layers, which means having fewer parameters to be tuned compared with the complete model. In this work, we propose to freeze the RNN part of the model and fine-tune only the final FC layer, with the insight that the RNN learns the features of the pHRI (*e.g.,* a force in a direction means that the human wants to steer the system in the same direction, an increasing force means the human is accelerating), while the FC layer learns how a specific user does interact (*e.g.,* how much force a particular user uses to steer the system, how fast a specific user accelerate the system, etc.). With the TL approach, it is possible to quickly adapt the model $\mathcal{M}_K$ to new users and additional objects co-carrying, with little data collection and fast training, leading to $\mathcal{M}_{TL} = \mathcal{M}_{TL}(\mathcal{M}_K, \mathcal{D}_K)$. The iterative procedure is summarized in Algorithm 1.

## III. EXPERIMENTS

The presented method is evaluated with real experiments. The robotic platform is a UR5 robot, with a Robotiq FT300 mounted at the tip for measuring the human interaction force. For this study, we want to simulate a collaborative motion along the x–y plane, which is typical for applications such as large object co-transportation. Moreover, the experiments want to simulate a situation in which a human and a robot are moving together along a trajectory, and the human, at a certain point, needs to deform the trajectory (e.g., because there is an obstacle that the robot doesn't know in between).

### A. Data collection

For the model's training, we collect the robot's actual poses and velocities from the robot's controller. The interaction force is measured at the robot tip via the FT sensor. An external computer computes the robot's nominal trajectories and streams the commands in real time. The data are sampled at 8 milliseconds. Three nominal robot trajectories are defined: linear, curved, and sinusoidal. The three trajectories are visible in Figure 3a, 3b, and 3c, respectively. The human must follow three trajectories visible on a monitor and deform them to avoid an obstacle that appears randomly at some point in the trajectory. The robot does not know the position of the obstacle.

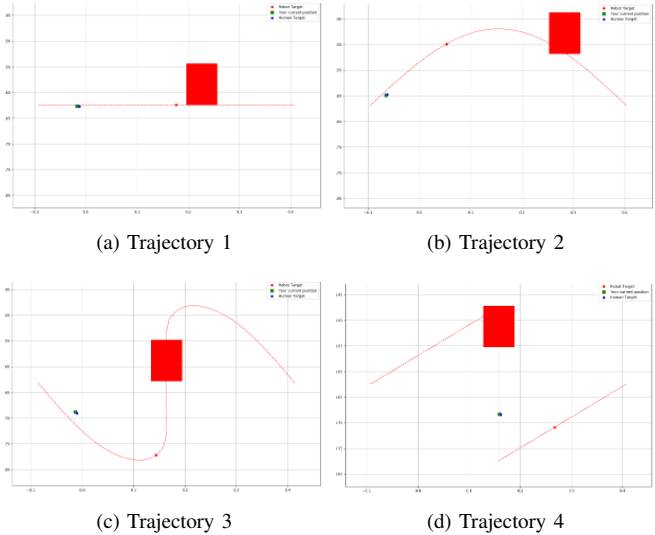

(a) Trajectory 1  (b) Trajectory 2
(c) Trajectory 3  (d) Trajectory 4

Fig. 3: The trajectories visible on the monitor. The red box is the obstacle, the green cross is the current position, and the red dot is the robot reference. The training uses trajectories 3a, 3b, and 3c. Trajectory 3d is for the evaluation and TL.

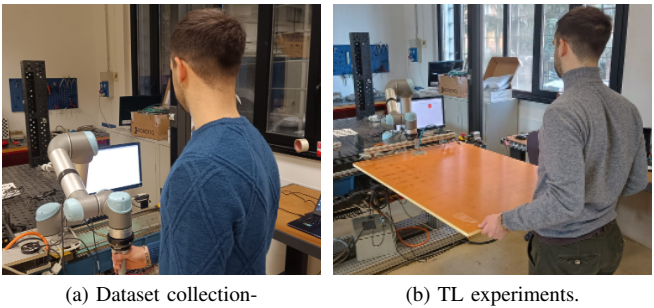

(a) Dataset collection-  (b) TL experiments.

Fig. 4: Experimental setup: a Robotiq FT300 sensor is at the robot tip; a monitor shows the reference trajectory.

To train the model, a single operator performed 20 repetitions for each trajectory for 60 trials. Despite the stop criterion (14) being matched after 1 iteration, we decided to perform 4 iterations to evaluate if iterating more can produce some improvement. The RNN+FC model has input data collected in the 125 time instant precedent, and the prediction horizon is set to 50 time steps. The LSTM model comprises 3 layers with 250 nodes, and the FC comprises two connected layers. The learning rate, decay function, and the optimizer of the RNN+FC model are obtained with Optuna [36]. The model is trained for 25 epochs.

The impedance parameters are $M_i = diag(10, 10)$, $C_i = diag(100, 100)$, and the stiffness is set to null $K_i = diag(0, 0)$, as typically in pHRI. The mass and damping coefficients have been hand-tuned to allow smooth motions. The cost of the two players are set as $Q_{h,h} = Q_{r,r} = diag([1, 1, 0.0001, 0.0001])$, $Q_{h,r} = Q_{r,h} = 0^{2\times2}$ and $R_h = R_r = diag([0.0005, 0.0005])$. The human cost function parameters $Q_{h,h}$, $Q_{h,r}$ and $R_h$ are recovered via Inverse Optimal Control (IOC) as in [37], and an average value is used. The robot parameters $Q_{r,r}$, $Q_{r,h}$, and $R_r$ are set equal to the human's to mimic a person. Different tuning may result in more assistive behavior, which might be desirable. Since the DGT controller accepts as a reference for (8) only one set-point rather than a prediction horizon, we take the $20^{th}$ point of the prediction horizon. Finally, $\alpha = 0.8$ is chosen to allow sufficient assistance. This value allows high assistance and the robot to recover the position of the robot set-point autonomously. Moreover, this value and higher allows the assumption $z_{ref} \simeq z_{ref,h}$ to hold. Figure 4 shows the setup.

*B. Model evaluation*

To evaluate the model's prediction capability, we measure the average RMS along each trajectory, computed as in (14). First of all, we evaluate the improvement provided by the iterative training. We compare the $e_{RMS}$ for four iterations. At each iteration $k$, data $\mathcal{D}_k$ are collected with the model trained on the previous iterations $\mathcal{M}_{k-1}$.

We also want to evaluate the error to the width of the prediction horizon. This information is relevant, for example, in a Model Predictive Control (MPC) implementation or to foresee in advance dangerous situations (*e.g.,* collisions that the human does not expect, proximity to the robot workspace boundaries, etc.). Therefore, we measure the $e_{RMS}$ with different prediction horizons. Note that we use the same model, so a horizon of 50 time steps is computed in all the cases, but we evaluate just the first $n$ samples of it for horizons of $H = \{5, 10, 20, 50\}$ time steps, which corresponds to 0.04, 0.08, 0.16, and 0.4 seconds. Figure 5 shows the results of the iterative training evaluations, also with the time horizon dependence. On the one hand, by iterating the model, the $e_{RMS}$ decreases significantly in two iterations, after that, remains stable, and no sensible improvements are visible after one iteration of the model. On the other hand, for a wide prediction horizon, the error increases. This is mainly because predicting human deviations from the nominal trajectory in advance is complex. Indeed, the $e_{RMS}$ increases just before the human starts the deviation from the nominal trajectory, and the robot cannot know it in advance. Despite this, the average error is about one millimeter and increases to about two with the maximum prediction horizon considered. Some improvement can be achieved in terms of prediction direction. This is visible qualitatively from Figure 6 and can be explained by evaluating the maximum prediction error. Therefore, the maximum expected error $e_{MAX}$ should be considered. Indeed, a significant error may cause an unwanted robot behavior that may cause it to go in the wrong direction because the human intention is misunderstood, even for some time instants. Results are visible in table I. It is visible that by iterating more, the

TABLE I: Model Evaluation: $e_{MAX}$ for different models $\mathcal{M}_k$ and horizons $H_k$. Values are in millimeters

|          | $\mathcal{M}_0$ | $\mathcal{M}_1$ | $\mathcal{M}_2$ | $\mathcal{M}_3$ |
|----------|-----------------|-----------------|-----------------|-----------------|
| $H_5$    | 11.1 ± 1.9      | 4.7 ± 3.5       | 4.1 ± 1.5       | 3.0 ± 0.4       |
| $H_{10}$ | 12.0 ± 1.8      | 5.0 ± 3.4       | 4.2 ± 1.5       | 3.2 ± 0.4       |
| $H_{20}$ | 14.3 ± 1.8      | 5.9 ± 3.3       | 4.9 ± 1.4       | 3.9 ± 0.5       |
| $H_{50}$ | 24.4 ± 2.9      | 13.7 ± 4.1      | 13.5 ± 2.2      | 10.6 ± 1.6      |

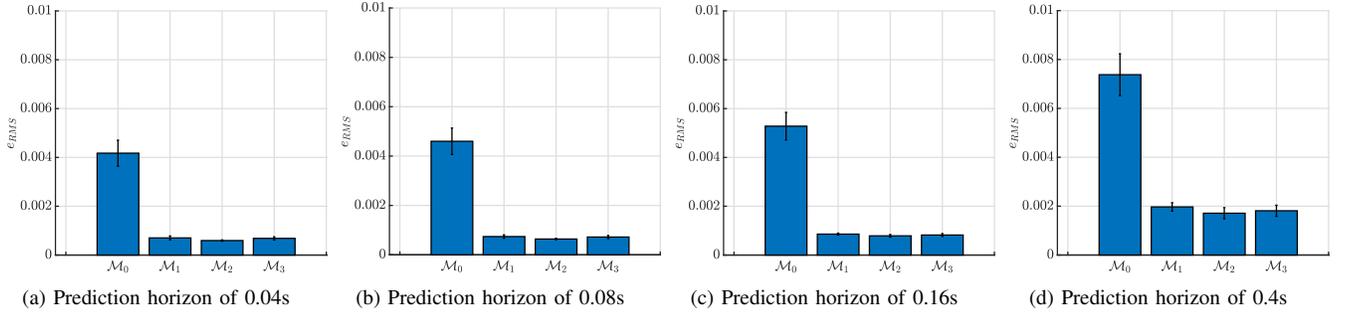

(a) Prediction horizon of 0.04s  (b) Prediction horizon of 0.08s  (c) Prediction horizon of 0.16s  (d) Prediction horizon of 0.4s

Fig. 5: Model Evaluation: $e_{RMS}$ for four models with different prediction horizons and model iterations $\mathcal{M}_k$.

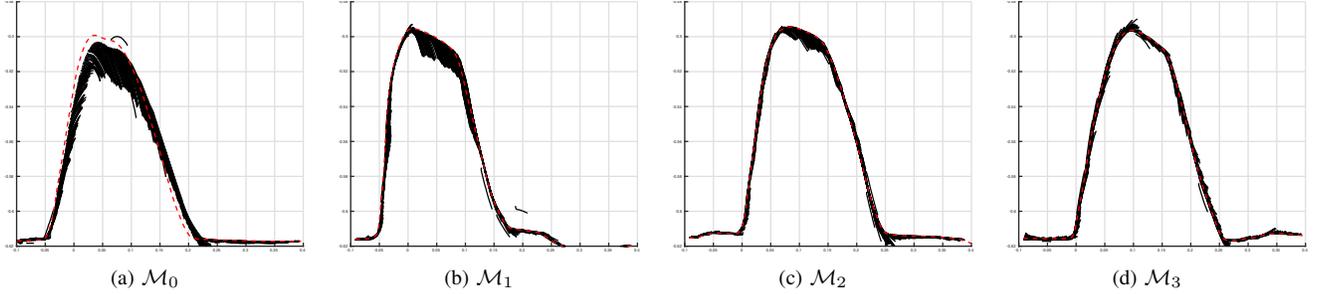

(a) $\mathcal{M}_0$  (b) $\mathcal{M}_1$  (c) $\mathcal{M}_2$  (d) $\mathcal{M}_3$

Fig. 6: Model Evaluation: the prediction at the various model iterations $\mathcal{M}_k$. The maximum prediction horizon is considered (0.4s). In solid black, the prediction at each time instant. In dashed red, the executed trajectory.

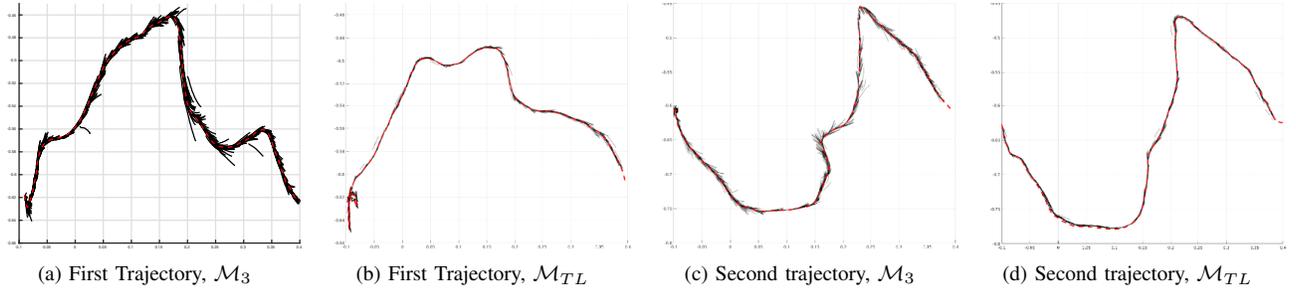

(a) First Trajectory, $\mathcal{M}_3$  (b) First Trajectory, $\mathcal{M}_{TL}$  (c) Second trajectory, $\mathcal{M}_3$  (d) Second trajectory, $\mathcal{M}_{TL}$ maximum error decreases. In general, both $e_{RMS}$ and $e_{MAX}$ should be considered in the evaluations. Indeed, even though the $e_{RMS}$ is similar, this might be because a significant error (but limited in time) "vanishes" if the $e_{RMS}$ is computed over a long trajectory where the average error is very low.

Fig. 7: Transfer Learning Evaluation, Case III: Tracking of two trajectories before and after TL. The considered maximum prediction horizon is $0.4s$. In solid black, the prediction at each time instant. In dashed red, the executed trajectory.

### C. Transfer Learning evaluation

*1) Case I:* We evaluate a trajectory not used during the training. First, we evaluate the model's capability to predict human intent with a new trajectory. Then, TL learns the new trajectory and compares the errors. As for the previous case, different prediction horizons are evaluated. Results are in table II. The $e_{RMS}$ is similar to the prediction made on the training trajectories, around 1 mm for small prediction horizons (up to $H_{20}$), and becomes larger for wider prediction horizons. The TL allows for reducing the error, making it comparable to that of the known trajectories. TL procedure improves the maximum error of the prediction, and its results are comparable to that of the known trajectories.

TABLE II: Transfer Learning Evaluation, Case I: $e_{RMS}$ and $e_{MAX}$ before and after TL, for different models $\mathcal{M}_k$ and horizons $H_k$. Values are in millimeters.

|   | $\mathcal{M}_3$ | | $\mathcal{M}_{TL}$ | |
|---|---|---|---|---|
|   | $e_{RMS}$ | $e_{MAX}$ | $e_{RMS}$ | $e_{MAX}$ |
| $H_5$ | 1.13 ± 0.034 | 9.60 ± 10.9 | 0.89 ± 0.049 | 3.15 ± 0.79 |
| $H_{10}$ | 1.15 ± 0.026 | 9.65 ± 10.8 | 0.95 ± 0.056 | 3.36 ± 0.60 |
| $H_{20}$ | 1.15 ± 0.017 | 9.96 ± 10.6 | 1.03 ± 0.060 | 3.62 ± 0.40 |
| $H_{50}$ | 2.71 ± 0.052 | 19.4 ± 5.43 | 1.68 ± 0.092 | 12.6 ± 2.54 |

TABLE III: Transfer Learning Evaluation; time required for the data collection and model training.

|   | Iterations | TL subjects | TL object |
|---|---|---|---|
| dataset collection | 60 ± 10 min | 5 ± 2 min | 5 ± 2 min |
| training | 45 ± 5 min | 4 ± 1 min | 4 ± 1 min |

*2) Case II:* Five subjects are asked to perform the three trajectories used during the training and deform them by directly grasping the robot at the tip.

*3) Case III:* The subject that trained the base model grasps a panel of 106x82cm assisted by the robot. By adding the object, different forces are exchanged. Figure 7 shows two trajectories before and after TL. Figure 8 shows the maximum prediction horizon considered (50 time steps, or 0.4 s) after and before applying the TL. The TL improves the predicting performances for the case of different subjects, making the average error after the TL comparable to the error of the iterated model, about 2mm. The TL decreases the errors also for the case of the co-manipulated object.

Table III shows the time required for each iteration's dataset collection and model training, comparing it with the TL time. The iterative procedure takes about two hours for each iteration. The TL approach allows performing this procedure only once. After that, the model can be adapted to new situations in about 10 minutes, thanks to the TL approach. The computation runs on a laptop with Intel i7 and NVIDIA GeForce 1050.

*D. Comparison with standard pHRI controllers*

We compare the proposed approach with two common strategies, Impedance Control (IMP) and Manual Guidance (MG). Both controllers rely on (1), with the difference that the robot assistive contribution is null ($u_r = 0$) for the IMP, and the stiffness matrix is null $K_i = 0^{6 \times 6}$ for MG. We maintain the same mass for the IMP case, superimpose the stiffness matrix to $K_i = diag([200, 200])$, and the damping to the 90% of the critical damping.

We compare the Root Mean Square of the interaction force, a common performance index for pHRI controllers,

$$f_{RMS} = \sqrt{\frac{1}{L} \sum_{k=0}^{L} f_k^2 \, dt}, \quad (15)$$

with $L$ length of the trajectory as before and $f_k$ the module of the measured interaction force at time instant $k$.

Figure 9 reports the comparison of $f_{RMS}$ with the other IMP and MG, considering our model after TL. The t-test shows no difference between MG and IMP, while between the TL and the other two, there is. Although the improvement might seem minor, with a different tuning of the controller (higher $\alpha$ or smaller $R_r$) the robot might increase the assistance so that the force decrease even more.

## IV. DISCUSSION

*1) Extension to 6dof robots:* We conducted preliminary tests to simulate a 3D pick-and-place task. For such an application, with the insight that 250 nodes work fine for the 2 DoFs case, we increased the hidden nodes of one half, *i.e.*, 125 nodes. Preliminary tests showed that the time for training the model increased linearly, with similar results. For a model considering 3 positions and 3 rotations, the training time should increase but remains limited, *e.g.*, 750 hidden nodes takes about 3 hours of training.

*2) Dataset acquisition:* We considered three trajectories. Despite this, the human had to deviate from the nominal trajectory to avoid the obstacle, so the model learned to predict deformations in each considered Cartesian direction, even in the backward direction. The application of the model to a new trajectory has proven feasible with comparable

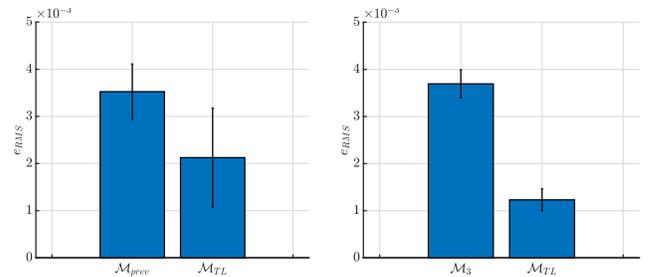

(a) Transfer to different subjects (Case II). 
(b) Transfer to different objects (Case III).

Fig. 8: Transfer learning Evaluation: value of $e_{rms}$

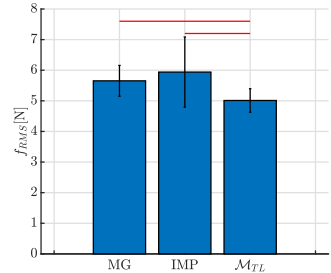

Fig. 9: The $f_{RMS}$ for Manual Guidance (MG), Impedance (IMP), and our assistive control with TL. A red bar indicates p-values $< 0.05$. The null hypothesis is rejected between $\mathcal{M}_{TL}$ and both MG and IMP, while it is not rejected between MG and IMP.

results. Despite the limitations of the current model, the method fits many applications. First, a human will unlikely need to use the full robot's workspace, while typically, humans work only in front of the robot. Second, applications require partially repetitive motions, such as pick and place. For example, consider composite material fabrication, where carbon fiber patches must be placed in layers on a mold. The patches are on a table and must be placed on the mold. The table and the mold always have the same position, and each patch's trajectory is similar, and co-manipulation is helpful for large patches. Defining a small set of trajectories is sufficient for such an application since the collaboration will always happen similarly and repetitively.

*3) Cooperative control:* We selected some parameters to tune the robot's level of assistance, such as $\alpha$ and $R_r$. Changing them, also the assistance that the robot can provide will vary. Applying the TL to fine-tune the model on humans and robots could be interesting and reasonable.

*4) Dataset dimension:* It is possible that better performances can be achieved with larger datasets, letting the model train for more than 25 epochs or other parameters. These aspects should be investigated in future works.

## V. CONCLUSION

This work presents a method for predicting the desired human trajectory over a finite time horizon. The model needs an iterative training procedure to reduce the prediction error, which is time-consuming and does not generalize to different users or situations. To overcome this issue, we propose a Transfer Learning method that adapts the model to new users and new situations. The approach is validated with real-world experiments. Different prediction horizons are also evaluated to show the dependency of the error. The time required by the different training steps is evaluated, showing that the TL approach reduces the time necessary to train the model. Finally, the assistive controller enhanced by the RNN+FC model is compared to standard controllers used in pHRI. Future works will focus on implementing an MPC integrating the full prediction horizon to compute the robot's assistive contribution. We will also address online modification of the assistance level and a model that considers the full 6 DoFs to co-manipulate deformable objects.

## ACKNOWLEDGMENT

This project has been funded by the European Union's Horizon 2020 research and innovation program under grant agreement No 101006732 (Drapebot).

This project has been partially funded by the project HYBRIDOpt, funded by Hasler Stiftung.